\documentclass[conference]{IEEEtran}
\IEEEoverridecommandlockouts
\usepackage{cite}
\usepackage{amsmath,amssymb,amsfonts}
\usepackage{algorithmic}
\usepackage{graphicx}
\usepackage{textcomp}
\usepackage{xcolor}
\def\BibTeX{{\rm B\kern-.05em{\sc i\kern-.025em b}\kern-.08em
    T\kern-.1667em\lower.7ex\hbox{E}\kern-.125emX}}

\makeatletter % changes the catcode of @ to 11
\newcommand{\linebreakand}{%
  \end{@IEEEauthorhalign}
  \hfill\mbox{}\par
  \mbox{}\hfill\begin{@IEEEauthorhalign}
}
\makeatother % changes the catcode of @ back to 12

\begin{document}

\title{Distinguishing Startle from Surprise Events Based on Physiological Signals %in MATB-II%Simulated Flight Scenarios
\\
% {\footnotesize \textsuperscript{*}Note: Sub-titles are not captured in Xplore and should not be used}
}

\author{
\IEEEauthorblockN{
Mansi Sharma\textsuperscript{1}, 
Alexandre Duchevet\textsuperscript{2}, 
Florian Daiber\textsuperscript{1}, 
Jean-Paul Imbert\textsuperscript{2}, 
Maurice Rekrut\textsuperscript{1}
}
\IEEEauthorblockA{
\textsuperscript{1} German Research Center for Artificial Intelligence (DFKI), \\ Cognitive Assistants Lab, Saarland Informatics Campus, Saarbrücken, Germany \\
Emails: \{Mansi.Sharma, Florian.Daiber, Maurice.Rekrut\}@dfki.de
}
\IEEEauthorblockA{
\textsuperscript{2} Fédération ENAC ISAE-SUPAERO ONERA, Université de Toulouse, France \\
Emails: \{alexandre.duchevet, Jean-Paul.IMBERT\}@enac.fr
}
}

\maketitle

% The abstract should state
% (1) the principal objectives and scope of the investigation,
% (2) describe the methods employed,
% (3) summarize the results, and
% (4) state the principal conclusions.

% \begin{abstract}
% This work aims at distinguishing startle from surprise events by classifying subtile changes in the physiological response based on measured biosignals, namely PPG, GSR, ECG, and Respiration Rate. We recorded data of xy participants during a controlled task in which we evoked startle, surprise and a combination of both events to trigger distinguishable physiological responses.
% \\
% use .... ML-methods? and fusion techniques...
% \\
% Our results show, that it is possible to distinguish startle from suprise and that a 

% \end{abstract}

\begin{abstract}
Unexpected events can impair attention and delay decision-making, posing serious safety risks in high-risk environments such as aviation. In particular, reactions like startle and surprise can impact pilot performance in different ways, yet are often hard to distinguish in practice. Existing research has largely studied these reactions separately, with limited focus on their combined effects or how to differentiate them using physiological data. %\Mansi{at this point, we should briefly about problems in existing studies: I am not aware of RW}
In this work, we address this gap by distinguishing between startle and surprise events based on physiological signals using machine learning and multi-modal fusion strategies.
% In this work, we conduct a controlled user study in an aviation-inspired task setting, where participants experienced Startle and Surprise events, while their physiological signals were recorded. 
% We conducted a controlled user study within an aviation-inspired task setting, where participants encountered Startle and Surprise events while their physiological signals were continuously recorded. 
Our results demonstrate that these events can be reliably predicted, achieving a highest mean accuracy of $85.7$\% with SVM and Late Fusion. To further validate the robustness of our model, we extended the evaluation to include a baseline condition, successfully differentiating between Startle, Surprise, and Baseline states with a highest mean accuracy of $74.9$\% with XGBoost and Late Fusion.

% \Mansi{briefly mention about our results}  
\end{abstract}

% Suggested rules for intro:
% (1) the introduction should present first, with all possible clarity,
% the nature and scope of the problem investigated.
% (2) it should briefly review the pertinent literature to orient the reader.
% (3) It should state the method of the investigation. If deemed
% necessary, the reasons for the choice of a particular method should be
% stated.
% (4) It should state the principal results of the investigation.
% (5) It should state the principal conclusions suggested by the results.
\section{Introduction}
The impact of startle and surprise on human cognitive performance has been a subject of interest, particularly within high-risk domains such as aviation \cite{metroStartle2024}. Unexpected events such as startle and surprise can disrupt cognitive processes, causing delays in decision-making and inappropriate actions that may have catastrophic consequences \cite{Dehais2015,Kinney2020}. 
%Such events, especially when involving both startle and surprise, are known to compromise performance by narrowing visual attention and degrading task accuracy of pilots \cite{Dehais2015,Kinney2020}. 
Previous research has identified that startle is a defensive reflex triggered by sudden and intense stimuli, producing physiological reactions such as increased heart rate, blood pressure, and skin conductance \cite{Koch1999,Gautier1997,Vrana1995}. This reaction can affect motor and cognitive tasks, with performance degradation spanning seconds to minutes \cite{woodhead1958effects,Thackray1983}. In contrast, surprise results from a discrepancy between expectation and reality, often requiring cognitive reframing to adjust to unexpected situations \cite{Klein2007}. Physiological indicators of surprise include increased skin conductance, heart rate deceleration, and pupil dilation \cite{Bradley2009,Reisenzein2012}.
% There are open datasets available for detecting surprise or startle effects, but to the best of our knowledge none specifically related to piloting activity. Recent work by \cite{duchevet2025} has investigated the effects of startle and surprise, either individually or combined, observed through behavioral and physiological measurements during the execution of a serious game simulating multiple tasks involved in piloting activity (NASA MATB-II  \cite{hart1988development}). The events were evoked while the participants were equipped with physiological sensors and performed a MATB-II task. 
While there are open datasets available for detecting surprise or startle effects, to the best of our knowledge, none focus specifically on piloting activity. Recent work by \cite{duchevet2025} addressed this gap by investigating startle and surprise—individually and combined—using behavioral (task performance, gaze) and physiological (heart rate, skin conductance) data collected during a serious game simulating piloting tasks (NASA MATB-II \cite{hart1988development}). Using repeated-measures ANOVA, they found that startle induced stronger physiological responses, while surprise mainly affected behavioral performance, with combined effects producing additive impairments.
\\
Traditional statistical techniques struggle with the complex interplay of physiological and behavioral data, limiting their ability to consistently differentiate startle from surprise and even in comparison with Baseline. We overcome this limitation by using machine learning to effectively distinguish between Startle, Surprise, Baseline, and their combinations. Our contributions include presenting the first machine learning approach to classify these states, achieving strong performance in pairwise comparisons and a three-class accuracy of $74.9$\%, which demonstrates the robustness of the method. Additionally, we introduce different fusion strategies both Early and Late fusion to combine physiological data and evaluate how varying temporal window durations ($3$s, $5$s, $7$s, and $10$s) impact classification performance.

% Our specific contributions are:
% \begin{enumerate}
%     \item We present the first machine learning approach to classify Startle, Surprise, and Baseline, demonstrating strong performance in pairwise comparisons as well as achieving a three-class accuracy of 74.96\%, highlighting the robustness of the method.
%     \item We present the first approach to use different fusion strategies (Early and Late) to combine physiological data. Subsequently, we evaluate on how temporal window durations ($3$s, $5$s, $7$s, and $10$s) affect performance. 
% \end{enumerate}
% The ability to differentiate these effects and automatically detect them based on physiological signals 
%in real time remains a significant challenge. 

% Therefore, the objective of this study is to use machine learning and multi-modal fusion strategies to distinguish between startle and surprise events based on physiological signals. 
% We evaluated ...
% Addressing this gap is essential for developing effective countermeasures that enhance safety in aviation and other safety-critical environments. 
% \Mansi{List of our Contributions}
% \begin{enumerate}
%     \item 
% \end{enumerate}

\begin{figure*}[t]
    \centering
    \includegraphics[width=\textwidth]{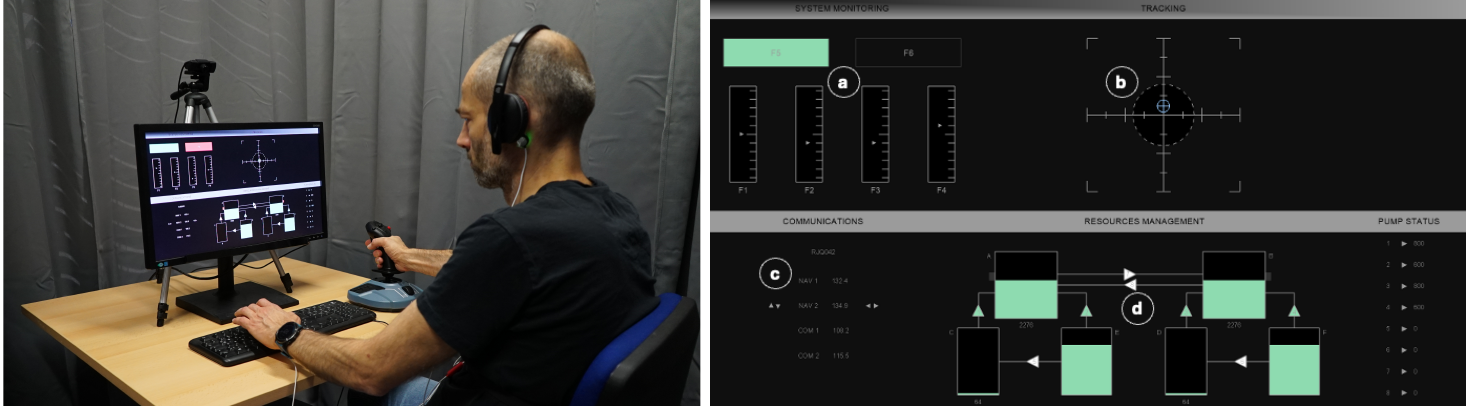}
    % [trim={left bottom right top},clip]
    \caption{Experimental set-up with the reverse video mode (surprising stimulus) of the Open-MATB (Right). Open-MATB interface with the four sub-tasks: system monitoring (a), tracking (b), communications (c), and resource management (d). %left: normal mode; right: reverse video mode (Left). %Figures are taken from the setup by \cite{duchevet2025}.
    } 
    \label{fig:study}
\end{figure*}

\section{Related Work}

%STARTLE

Startle is a defensive reflex triggered by an adverse event ~\cite{koch1999b} Its effect is characterized by a rapid and involuntary response to a sudden stimulus~\cite{koch1999b}. This stimulus can be either tactile~\cite{hoffman1964startle}, vestibular~\cite{bisdorff1995responses}, visual~\cite{bradley1990emotion}, or acoustic~\cite{yeomans1995acoustic}. A startle event triggers a rapid reaction, as quick as $53$ms for behaviors like eye closure~\cite{ekman1985startle}. 
The body startle response is induced by the activation of the sympathetic nervous system which is part of the body's broader "fight or flight" mechanism~\cite{cannon1932wisdom}, a concept widely discussed in the literature on physiology and psychology~\cite{mccarty2016}, and explain how an individual reacts to perceived threats.
%Ekman also noted that activities like neck muscle contractions and trunk movements occur within 200 ms. Physiological changes include increased heart rate peaking 4 s after the stimulus~\cite{gautier1997relationships}, higher blood pressure~\cite{holand1999effects}, and increased skin conductance~\cite{vrana1995emotional}.
Motor tasks are significantly impacted post startle. Tracking tasks are impaired for the first $2$s, with performance normalizing after up to $10$s~\cite{may1971, thackray1970, vlasak1969effect}. 
Cognitive tasks suffer similarly, with performance on visual matching tasks deteriorating for up to $31$s after a startling stimulus~\cite{woodhead1958effects}. Startle also decreases the performance of tasks requiring sustained mental effort, such as continual
mental subtraction~\cite{vlasak1969effect} and simulated air traffic control tasks~\cite{Thackray1983}. Interestingly, the startle reaction can be enhanced or attenuated by the emotional state of the individual~\cite{bradley2000, Lang1990, vrana1995emotional}.

%Surprise
Surprise is an emotion arising from a discrepancy between expectations and reality~\cite{Horstmann2006}. It can result from an unexpected or absence of expected change and is not necessarily accompanied by startle. A surprising event is sometimes marked by a universal facial expression~\cite{ekman1969pan, Hiatt1979}. Physiological responses to surprise include increased skin conductance~\cite{Bradley2009}, heart rate deceleration~\cite{Reisenzein2012}, and pupil dilation~\cite{Maher1979}. Surprise provokes a shift of attention to the surprising stimulus, facilitating the evaluation of the discrepancy, but interrupting ongoing activities~\cite{Tomkins1962}. Horstmann~\cite{Horstmann2006} found that this interruption lasts almost $1$s for $78$\% of participants. Making sense of a surprising situation requires updating or elaborating a new mental model~\cite{Klein2007}, a process that can delay reactions. Meyer et al.~\cite{Meyer1991} reported increased reaction times following visual surprises, with differences of up to $700$ms between experimental and control groups.

% TODO MORE ML!!??
Machine Learning has been used to measure cognitive states like channelized attention, diverted attention, including startle/surprise in pilots using physiological signals, create models assisting in training, decision making, and flight safety~\cite{Harivel2017, Harivel2016}. Authors used machine learning models like Random Forest, SVM but classification of highly overlapping states like startle and surprise remained challenging. In a follow-up work by \cite{Harivel2017}, authors focused on detecting attentional and stress-related states in pilots with real-time physiological data monitoring to support cockpit feedback and detected stress and workload, but did not separate startle from surprise, showing the need for more fine-grained models.  
Duchevet et al. \cite{duchevet2025} recently conducted a controlled MATB-II study, inducing startle and surprise through visual and auditory stimuli while recording physiological responses. Although they identified statistical differences between the two reactions, no machine learning methods were used for classification. In a more applied setting, Vlaskamp et al. \cite{vlaskamp2025recovery} conducted a large-scale survey involving $239$ airline pilots to assess operational strategies for recovering from startle and surprise events. They introduced the Reset Method, a structured behavioral response to regain situational awareness. While this study emphasizes the importance of managing such reactions, it does not involve physiological sensing or computational prediction.
\\
Although foundational work has shown distinct physiological responses to startle and surprise, few have used machine learning to reliably differentiate these reactions, especially against baseline. Our work addresses this gap by applying machine learning with multi-modal data fusion to classify startle, surprise, and baseline states.

\section{Data Recording}
Building on the previous work, we used the dataset from \cite{duchevet2025} which was collected in a controlled aviation-inspired setting using the NASA MATB-II task~\cite{duchevet2025}. 
\paragraph{Participants} From the dataset, we used a subset with 34 participants that included trials labeled as Startle and Surprise events, with each participant experiencing only one of these events per trial. The participants were aged between $21$ and $45$ years and had no prior flight experience.
% from $34$ participants where the data quality was good enough for a machine learning dataset. 

% (mean = $28.1$, SD = $6.5$). \Mansi{can someone check for mean sd, I do no have this age of all participants}
% All participants had at least a high school diploma, and no prior flight experience was necessary for the experimental task. 
\paragraph{Hardware Setup}  
Participants were equipped with over-ear headphones, a standard keyboard, a mouse, and a joystick to interact with the MATB-II task, displayed on a 19-inch monitor. Physiological signals were recorded using the Bitalino device\footnote{https://www.downloads.plux.info/OpenSignals/OpenSignals-Manual.pdf}, which included a three-lead electrocardiogram (ECG) to monitor heart rate, a galvanic skin response sensor for measuring skin conductance (EDA), a photoplethysmogram (PPG) earpiece to assess peripheral blood flow, and a respiration sensor belt placed around the chest to capture breathing patterns (RESP). LabStreamingLayer \footnote{https://github.com/sccn/labstreaminglayer} software was used for data recording and synchronization.

\paragraph{Recording procedure} 
Authors conducted the experiments using Open MATB software~\cite{cegarra2020openmatb}, an open-source implementation of the classic MATB-II task, which simulates multiple task management on a flight deck. The task included four sub-tasks: system monitoring, tracking, communications, and resource management (see Figure~\ref{fig:study} (right)). First the participants were asked to perform a ~$5$-minute tutorial session to get used to the MATB task itself. 
%The actual scenario lasted $7$ minutes, with a baseline period ($3$:$30$-$3$:$45$) and a study period ($5$:$00$-$5$:$15$) to assess post-stimulus performance. 
% Task difficulty was adjusted to ensure it was challenging but manageable. 
% Afterwards, participants performed a single $7$-minute experimental scenario under one of two conditions: startle, surprise.
Participants then completed a single $7$-minute experimental scenario under either the startle or surprise condition.
%, or a combination of both.
To induce the startle condition, participants were exposed to a $100$ dB startling noise at time $5$:$00$. Prior to the startle noise, three $80$ dB \textit{startle decoys} were evoked during the communication task for familiarization with noise effects. Although participants were informed about the potential sound but its exact timing was not disclosed. For the surprise condition, a reverse video effect was applied at time $5$:$00$ to provoke surprise, lasting until the end of the scenario. Figure \ref{fig:study} describes the experimental design from \cite{duchevet2025}, where a participant is engaged in the resource management task.
% In the combination condition, participants experienced both, the $100$ dB startling noise and the reverse video effect at $5$:$00$. \Mansi{Study setup image}
%In addition to physiological and eye tracking measures, we recorded across the sub-tasks performance metrics like reaction times and accuracy. 

% \begin{figure*}[t]
%      \includegraphics[
%      width=1\linewidth]{pics/Bestresults.pdf}
%      \caption{Mean Accuracy for Startle vs. Baseline with best performing $7$s window (left), Surprise vs. Baseline with best performing $5$s window (middle) and Startle vs. Surprise with best performing $5$s window (right).}
%      \label{fig:bestResults}
% \end{figure*}

\begin{figure*}[t]
     \includegraphics[
     width=1\linewidth]{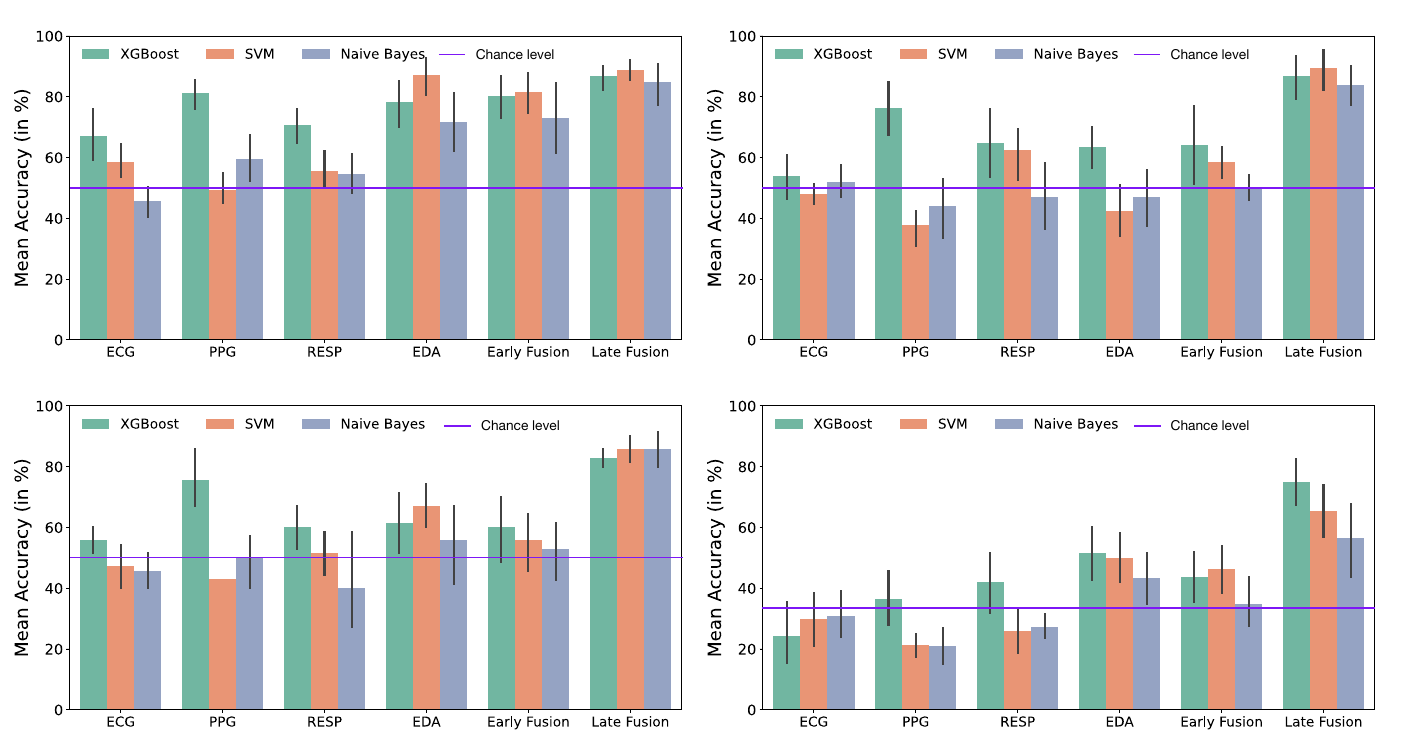}
     \caption{Mean Accuracy for \textbf{Startle vs Baseline} with best performing $3$s window (top left, chance level of $50\%$), \textbf{Surprise vs Baseline} with best performing $7$s window (top right, chance level of $50\%$), \textbf{Startle vs Surprise} with best performing $10$s window (bottom left, chance level of $50\%$), and \textbf{Startle vs Surprise vs Baseline} with best performing $3$s window (bottom right, chance level of $33.3\%$). Error bars indicate $95$\% confidence interval estimated via bootstrapping around the mean.}
     \label{fig:bestResults}
\end{figure*}

\section{Methods}
The following sections describes the pre-processing and feature extraction procedures for the physiological data, followed by an overview of our classification and data fusion methods.

\paragraph{Pre-processing and Feature extraction}
% The bio-signal data analysis involved an offline analysis of the recorded data including a structured preprocessing pipeline to ensure noise-free and reliable feature extraction. ECG signals were processed using a high-pass filter ($0.6$ Hz), a low-pass filter ($100$ Hz), and a notch filter ($50$ Hz) to remove baseline drift and power line interference. PPG signals were band-pass ($0.5$–$5$ Hz) and high-pass ($0.5$ Hz) filtered to retain relevant frequency components. RESP and EDA signals were smoothed using a low-pass filter with a $5$ Hz cutoff frequency. 
% Feature extraction included statistical features such as mean, standard deviation, minimum, and maximum values, alongside peak count, which was determined by the number of signal peaks exceeding the mean amplitude.

The bio-signal data analysis was performed offline, enabling detailed evaluation of the recorded raw physiological signals. A structured pre-processing pipeline inspired from previous work \cite{alreshidi2024advancing, Harivel2016, metroStartle2024} was implemented to ensure the removal of artifacts and preservation of relevant physiological information. For ECG, a $4$th-order Butterworth high-pass filter (cutoff:$0.6$ Hz, 24 dB/octave) was applied to eliminate baseline drift caused by slow physiological processes such as respiration. To suppress high-frequency noise, a low-pass Butterworth filter with a $100$ Hz cutoff was used. Additionally, a notch filter centered at $50$ Hz was used to remove power line interference. PPG signals were pre-processed using a 4th-order Butterworth band-pass filter ranging from $0.5$ to $5$ Hz, along with a high-pass filter at $0.5$ Hz, to retain frequency components relevant to pulse wave activity while discarding extraneous noise. RESP and EDA  were smoothed using a low-pass Butterworth filter with a $5$ Hz cutoff, reducing high-frequency noise while preserving the slower variations characteristic of these modalities. Following pre-processing, features were extracted from the cleaned signal modalities.
Feature extraction involved computing a range of statistical descriptors that capture key characteristics of the signal. These included features such as the mean, standard deviation, minimum, and maximum values. This provides insight into the overall distribution of the variability of the signal over time. 
%In addition, we computed the number of peaks exceeding the mean amplitude-referred to as the peak count, which was calculated to quantify the frequency of prominent signal fluctuations. 
In addition, we computed the number of local maxima (peaks) whose amplitude exceeded the mean signal amplitude-referred to as the peak count, to quantify the frequency of prominent signal fluctuations.

\paragraph{Classification and Fusion Techniques}
For classification, we used three machine learning models, Support Vector Machine (SVM), Na{\"i}ve Bayes (NB), and XGBoost as a popular choice in handling physiological data \cite{alreshidi2024advancing, metroStartle2024}. To leverage the complementary information across different physiological modalities, we adopted both early and late fusion strategies. Previous research with multi-modal data has shown significant improvement over uni-modal performance and has enhanced the accuracy and generalizability in various domains including aviation \cite{Harivel2017, sharma2023implicit, sharma2024distinguishing}. Inspired by this, we implemented two different fusion mechanisms. In the early fusion approach, features extracted from multiple bio-signals (i.e., ECG, PPG, EDA, RESP) were concatenated into a single feature vector before being input into the classifiers, allowing the models to learn joint patterns across modalities. In addition, we implemented a late fusion technique based on majority voting, where individual classifiers were trained separately on each modality, and the final prediction was determined by aggregating their outputs through a majority vote. This ensemble method helped enhance robustness by combining independent decisions from each uni-modal classifier, reducing the influence of modality-specific noise or misclassifications.
\\
Startle and Surprise events were cut from the data according to the annotation. For the Baseline, we selected a window in the signal prior to the onset of these events to ensure that Startle and Surprise effects do not influence the Baseline. Overall, we had $17$ samples for each class.\\
We followed a standard procedure for model evaluation with hyperparameter tuning using $5$-fold cross-validation to ensure robust performance. 
%The dataset was split into training and testing subsets using an $80$:$20$ ratio. 
Each sample contained only one condition, and the dataset was split at the sample level into training and testing sets with $80$:$20$ ratio with no overlap between splits.
Additionally, different window durations ($3$s, $5$s, $7$s, and $10$s) were explored to assess their influence on classification accuracy as inspired by the state-of-the-art \cite{rivera2014startle, metroStartle2024}.

\section{Results}
\sloppy
In the following, we present our classification results for four evaluation experiments: Startle vs Surprise, and three additional comparisons involving the Baseline condition, i.e., Startle vs Baseline, Surprise vs Baseline, and Startle vs Surprise vs Baseline. We also analyze the impact of varying window sizes across these experiments.

% In the Startle vs. Baseline condition, XGBoost with Early Fusion achieved highest accuracy of $88.64\%$. For Surprise vs. Baseline and Startle vs. Surprise, SVM with Early Fusion reached highest accuracy of $80.49\%$ and $72.86\%$, respectively. While for the Startle vs Baseline condition $7$s emerged as the best time window, the Surprise vs Baseline and Startle vs Surprise condition achieved best results with the $5$s window.
%In the Startle vs Surprise condition, SVM slightly surpassed XGB with the early fusion and the $5$s time window.
% Between conditions we found EDA to be a reliable feature with early fusion leading to significant improvement in the Startle vs Surprise condition. Further, the $3$s time window lead to lower performance in all conditions. This suggests that a moderately long window balances feature richness and computational efficiency while short windows (e.g., $3$s) leading to lower accuracy, and excessively long windows (e.g., $10$s) may introduce redundant or noisy data, slightly reducing performance compared to the optimal $5-7$s window.

\subsubsection{Startle vs Surprise}
Figure \ref{fig:bestResults} (bottom left) demonstrates the classification result in distinguishing Startle vs Surprise events across different physiological signals and fusion approaches with best performing $10$s window. Late fusion achieved significantly high accuracy with mean accuracy of $85.7\%$ for both SVM and Na{\"i}ve Bayes compared to uni-modal approaches and early fusion, indicating the effectiveness of our fusion method. Notably, XGBoost exhibited strong performance with individual modalities in this window, particularly PPG (mean accuracy of $75.7$\%) and ECG (mean accuracy of $55.7$\%), suggesting its effectiveness in leveraging  physiological features for longer-duration event detection.
\\
For shorter time windows ($3$s, $5$s, and $7$s), late fusion consistently surpassed uni-modal approaches, demonstrating again its robustness in integrating multi-modal physiological data. Interestingly, EDA consistently outperformed other individual modalities across most window durations, achieving its highest mean accuracy of $78.6$\% with a $7$s window. For the $10$s window, however, PPG outperformed all other modalities with a mean accuracy of $75.7$\%.

\subsubsection{Comparison with Baseline} We outline our evaluation experiments aimed at predicting Startle versus Surprise events in relation to the Baseline resulting in three evaluations: Startle vs Baseline, Surprise vs Baseline, and Startle vs Surprise vs Baseline.
Figure \ref{fig:bestResults} (top left) present the classification of Startle vs Baseline states using the aforementioned physiological signals and with performing window sizes i.e., $3$s. SVM with late fusion yielded the highest accuracy of $88.8$\%. Among individual modalities, EDA proved to be the most reliable for distinguishing Startle from Baseline, with SVM achieving a mean accuracy of approximately $87.0\%$. This finding is consistent with the classification results of distinguishing Startle vs Surprise event, where EDA also performed better than other modalities. Late fusion improved the accuracy but not significantly compared to EDA alone for this specific window. PPG alone achieved good results with XGBoost model. 

For Surprise vs Baseline comparison as in Figure \ref{fig:bestResults} (top right), SVM achieved the highest mean accuracy of $89.6$\% with late fusion in $7$s window. For other uni-modalities, XGBoost performed better than the other classifiers. In this evaluation, PPG achieved higher performance compared to other uni-modality. 
To ensure robustness of our classification methods in predicting these events with Baseline, we evaluated on differentiating Startle vs Surprise vs Baseline, where late fusion with XGBoost 
significantly performed better than other uni-modal and early fusion approaches with achieving highest accuracy of $74.9$\%. Na{\"i}ve Bayes classifier did not achieved better results compared to other models in all Baseline evaluations. Overall, our results revealed that our methods proved to be effective and robust in predicting Startle, Surprise, and Baseline events. 

% Na{"i}ve Bayes classifier not good in all abseline.

% \Mansi{maybe a visualization of windows for best classifier for each condition would be good, describe figure 3}

\begin{figure*}[t]
     \includegraphics[
     width=1\linewidth]{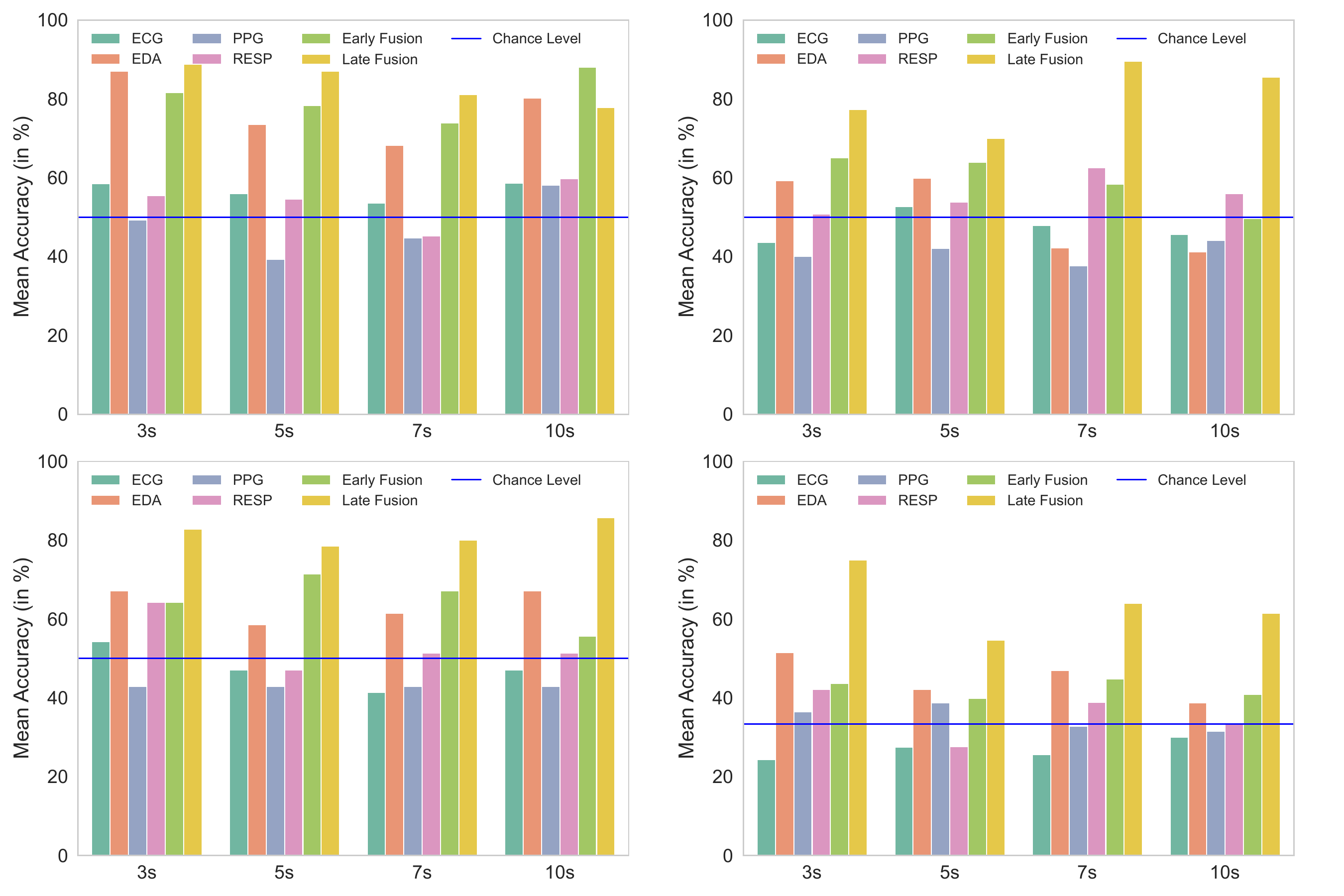}
     \caption{Mean accuracy across all time windows is shown for each classification task: \textbf{Startle vs Baseline} using the best-performing SVM model (top left, chance level of $50\%$), \textbf{Surprise vs Baseline} using the best-performing SVM model (top right, chance level of $50\%$) , \textbf{Startle vs Surprise, chance level of $50\%$} using the best-performing SVM model (bottom left, chance level of $50\%$), and \textbf{Startle vs Surprise vs Baseline} using the best-performing XGBoost model (bottom right, chance level of $33.3\%$).}
     \label{fig:allwindows_bestmodel}
\end{figure*}

\subsection{Evaluation across window sizes}
To illustrate the effect of window sizes on classification performance, we present a visualization in Figure \ref{fig:allwindows_bestmodel} by utilizing the best performing classifier i.e., for binary classification experiments SVM achieved the best results and with three-class classification, XGBoost achieved the best results. 

Across all conditions and window durations, the late fusion approach consistently outperformed models based on individual modalities, highlighting the benefit of multi-modal integration.
In the Startle vs Baseline condition (Figure \ref{fig:allwindows_bestmodel}, top left), the classification performance remained largely unaffected by the choice of window size. All tested durations yielded similar accuracy, suggesting that the distinguishing features between these two states are relatively stable over time.
A comparable pattern was observed in the Startle vs Surprise condition (Figure \ref{fig:allwindows_bestmodel}, bottom left), where performance was also robust across different window lengths. This indicates that the temporal dynamics of startle and surprise states may be equally captured across short and long observation periods. However, in the Surprise vs Baseline comparison (Figure \ref{fig:allwindows_bestmodel}, top right), a different trend emerged, longer time windows, specifically $5$s and $7$s consistently resulted in better classification performance than shorter ones. This suggests that capturing the full temporal evolution of surprise may require a longer observation window to differentiate it reliably from the baseline state. Interestingly, in the Startle vs Surprise vs Baseline three-class classification task (Figure \ref{fig:allwindows_bestmodel}, bottom right), the shortest window duration of $3$s yielded the highest accuracy, surpassing the performance of all longer windows. This finding implies that the early-stage features within the initial few seconds may be most discriminative when distinguishing among all three affective states simultaneously.

% EDA modality, reaching approximately $70$\%, showing the effectiveness of EDA in distinguishing surprise from Baseline. On the other hand, Naive Bayes achieves the lowest mean accuracy with the PPG modality, around $40$\%, which is similar performance compare to Startle vs. Baseline. Overall, XGBoost outperforms SVM and Na{\"i}ve Bayes across all modalities, with Early Fusion enhancing accuracy for all models. The $10$s window achieves the highest accuracy overall, particularly for XGBoost with EDA and Early Fusion modalities. This suggests that longer windows capture more meaningful physiological features for distinguishing surprise from Baseline. Shorter windows (e.g., $3$s) result in lower accuracy due to insufficient data for robust feature extraction, similar to Startle vs. Baseline. As the window size increases, accuracy improves across all models, with Early Fusion benefiting the most from longer windows.

%{Startle vs Surprise vs Baseline}

\section{Discussion}
By integrating ECG, EDA, RESP, and PPG signals through a structured pre-processing pipeline and extracting features from raw physiological data, our classification framework demonstrated that machine learning techniques, particularly late fusion, can effectively distinguish between Baseline, Startle, and Surprise conditions. Specifically, late fusion consistently outperformed uni-modal approaches, highlighting the complementary nature of different physiological signals in capturing cognitive-affective states. Among the modalities, EDA and PPG features proved most informative, contributing significantly to the overall classification performance. Our results showed high accuracy across multiple binary comparisons, with the SVM model achieving up to $89.6$\% accuracy in distinguishing Surprise from Baseline conditions. Even in the more complex three-class scenario, the XGBoost model achieved $74.9$\% accuracy, more than twice the chance level, demonstrating the robustness of the proposed approach. Window size also played a crucial role in model performance. For binary classifications such as Startle vs Baseline or Surprise vs Startle, performance remained relatively stable across different window lengths. However, in the Surprise vs Baseline , longer windows ($5$s and $7$s) provided a clearer physiological distinction, whereas in the more dynamic three-class task, shorter windows ($3$s) led to significantly better performance. These findings suggest that task complexity and the nature of the physiological response should inform the choice of temporal windowing in real-time applications.

\section{Conclusion and Future Work}
Unexpected events such as Startle and Surprise significantly impair attention and delay decision-making, especially in high-risk environments like aviation. These reactions differ physiologically and cognitively but are often difficult to distinguish in practice. Our study addressed this gap by conducting a controlled aviation-inspired task where participants encountered startle and surprise events while physiological signals were recorded. We demonstrated that startle and surprise can be reliably classified using multi-modal physiological data, achieving up to $85.7$\% accuracy in differentiating these events and $74.9$\% accuracy when including a Baseline condition. This work highlights the feasibility of enhancing safety by timely detection of cognitive disruptions caused by unexpected events. Current approaches rely on selecting controlled stimuli and limited datasets, which is valuable for preliminary research. 
Future work should expand datasets to include more naturalistic, combined startle-surprise scenarios, improve model generalization across individuals, and integrate detection systems into adaptive cockpit tools.
\section{Acknowledgements}
This project has received funding from the European Union’s Horizon Europe research and innovation programme HORIZON-CL5-2021-D6-01-13 under Grant Agreement no 101075332.

% The preferred spelling of the word ``acknowledgment'' in America is without 
% an ``e'' after the ``g''. Avoid the stilted expression ``one of us (R. B. 
% G.) thanks $\ldots$''. Instead, try ``R. B. G. thanks$\ldots$''. Put sponsor 
% acknowledgments in the unnumbered footnote on the first page.

%\section*{References}
%\begin{thebibliography}{00}
\bibliographystyle{plain}
\bibliography{refs}
% Please number citations consecutively within brackets \cite{b1}. The 
% sentence punctuation follows the bracket \cite{b2}. Refer simply to the reference 
% number, as in \cite{b3}---do not use ``Ref. \cite{b3}'' or ``reference \cite{b3}'' except at 
% the beginning of a sentence: ``Reference \cite{b3} was the first $\ldots$''

% Number footnotes separately in superscripts. Place the actual footnote at 
% the bottom of the column in which it was cited. Do not put footnotes in the 
% abstract or reference list. Use letters for table footnotes.

% Unless there are six authors or more give all authors' names; do not use 
% ``et al.''. Papers that have not been published, even if they have been 
% submitted for publication, should be cited as ``unpublished'' \cite{b4}. Papers 
% that have been accepted for publication should be cited as ``in press'' \cite{b5}. 
% Capitalize only the first word in a paper title, except for proper nouns and 
% element symbols.

% For papers published in translation journals, please give the English 
% citation first, followed by the original foreign-language citation \cite{b6}.

% \vspace{12pt}
% \color{red}
% IEEE conference templates contain guidance text for composing and formatting conference papers. Please ensure that all template text is removed from your conference paper prior to submission to the conference. Failure to remove the template text from your paper may result in your paper not being published.

\end{document}